\documentclass[10pt,twocolumn,letterpaper]{article}
\pdfoutput=1
\usepackage[final]{cvpr} 

\usepackage{hyperref}       % hyperlinks
\usepackage{url}            % simple URL typesetting
\usepackage{multirow}
\usepackage{colortbl}

\hypersetup{
	colorlinks=true,
	citecolor=teal,
}

\definecolor{cvprblue}{rgb}{0.21,0.49,0.74}
\def\paperID{9242} % *** Enter the Paper ID here
\def\confName{CVPR}
\def\confYear{2025}

\title{Multi-Stage Vision Token Dropping: Towards Efficient \\Multimodal Large Language Model}

\author{Ting Liu\textsuperscript{1}\thanks{Equal contribution.} \ \ Liangtao Shi\textsuperscript{2}\footnotemark[1] \ \ Richang Hong\textsuperscript{2} \ \ Yue Hu\textsuperscript{1} \ \ Quanjun Yin\textsuperscript{1}\textsuperscript{$\dagger$} \ \ Linfeng Zhang\textsuperscript{3}%\textsuperscript{$\ddagger$}
\thanks{Corresponding authors}
\\
  \textsuperscript{1}National University of Defense Technology\ \ \textsuperscript{2}Hefei University of Technology\ \  \\
  \textsuperscript{3}Shanghai Jiao Tong University\\
  \normalsize
  \texttt{liuting20@nudt.edu.cn, shilt@mail.hfut.edu.cn, zhanglinfeng@sjtu.edu.cn} \\}
  
\begin{document}
\maketitle
\begin{abstract}

The vision tokens in multimodal large language models usually exhibit significant spatial and temporal redundancy and take up most of the input tokens, which harms their inference efficiency. To solve this problem, some recent works were introduced to drop the unimportant tokens during inference where the importance of each token is decided only by the information in either the vision encoding stage or the prefilling stage. In this paper, we propose Multi-stage Token Dropping (MustDrop) to measure the importance of each token from the whole lifecycle, including the vision encoding stage, prefilling stage, and decoding stage.
Concretely, in the visual encoding stage, MustDrop merges spatially adjacent tokens with high similarity, and establishes a key token set to retain the most vision-critical tokens, preventing them from being discarded in later stages. In the prefilling stage, MustDrop further compresses vision tokens by the guidance of text semantics, with a dual-attention filtering strategy. In the decoding stage, an output-aware cache policy is proposed to further reduce the size of the KV cache. By leveraging tailored strategies in the multi-stage process, MustDrop can more precisely recognize the important and redundant tokens, thus achieving an optimal balance between performance and efficiency. For instance, MustDrop reduces about 88.5\% FLOPs on LLaVA with a compression ratio of 92.2\% while maintaining comparable accuracy. Our codes are available at \url{https://github.com/liuting20/MustDrop}.

\end{abstract}    
\begin{figure}[t]
\centering
\includegraphics[width=1.0\columnwidth]{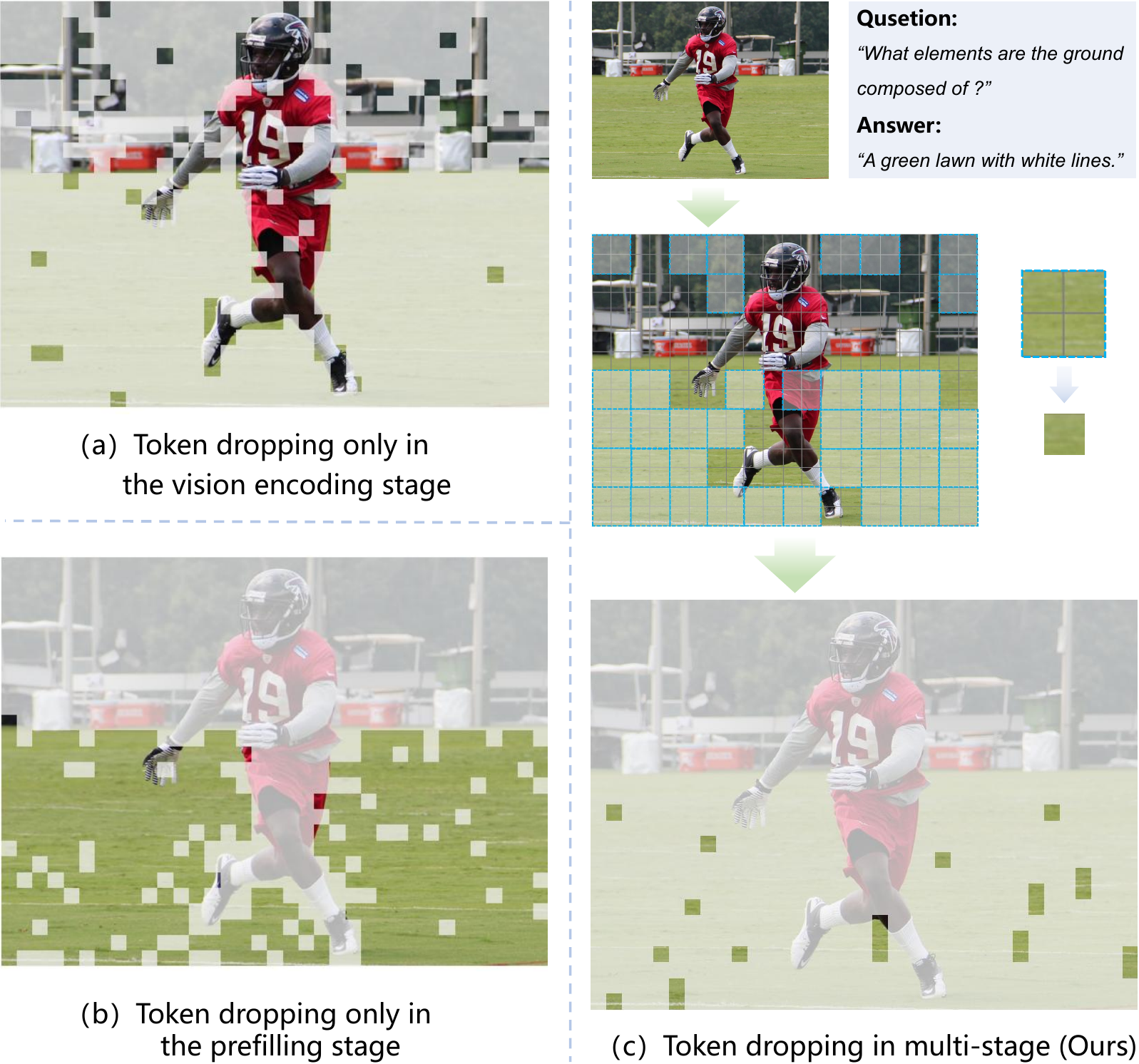}
\caption{Comparison of vision token dropping methods: (a) methods that only drop tokens during the vision encoding stage, i.e., PruMerge and ToMe, (b) methods that remove tokens limited to the prefilling phase, i.e., FastV and SparseVLM, and (c) our Mustdrop approach, which gradually removes invalid tokens during the vision encoding, prefilling, and decoding stages.
}
\label{fig:intro}
\end{figure}

\section{Introduction}
\label{sec:intro}
Following the tremendous advancements in Large Language Models (LLMs) \citep{brown2020language, achiam2023gpt, touvron2023llama, glm2024chatglm,dettmers2024qlora,rafailov2024direct}, Multimodal Large Language Models (MLLMs) \citep{team2023gemini, Qwen-VL,li2023blip,xiao2024can,li2024univs} subsequently endow LLMs with the ability of vision perception. The mainstream MLLMs first process images by a vision encoder, and then input vision tokens together with text tokens into the LLM for multimodal interaction. The rapid development of MLLMs facilitates the convenience that artificial intelligence achievements \citep{hong2024cogagent,you2025ferret,huang2025lita} bring to daily life. 

Despite the promising advancements in MLLMs, a surge of visual tokens inevitably leads to huge memory and computational costs \citep{ju2025turbo}. 
The root of this problem lies in the design of the Transformer \citep{vaswani2017attention} architecture, in which the computation costs tend to increase quadratically with the length of input tokens. Obviously, compared with text tokens, visual tokens possess a significant amount of redundancy since the vision tokens which have close temporal and spatial positions usually exhibit similar information, in other words, redundant information. The above problem is more pronounced for high-resolution images and long videos. To address this problem, some works \citep{shang2024llava,arif2024hired,li2023blip,yao2024deco} focus on removing tokens in the vision encoding phase but fail in determining the related-text visual tokens, and blindly deleting them will inevitably damage the multimodal understanding. Considering the importance of text semantics, FastV \citep{chen2024image} and SparseVLM \citep{zhang2024sparsevlm} are proposed for removing irrelevant-text vision tokens, which highlights the interaction between multi-modality but ignore the redundant and crucial semantics in the vision modality.

These problems raise an intuitive question - \emph{``how to select the most suitable tokens for pruning based on the information from the whole inference process of MLLM?''.
}
In this paper, we answer this question by first delving deeply into the roles of the three computation stages in MLLM, including vision encoding, prefilling, and decoding.

\begin{figure*}[t]
\centering
\includegraphics[width=\textwidth]{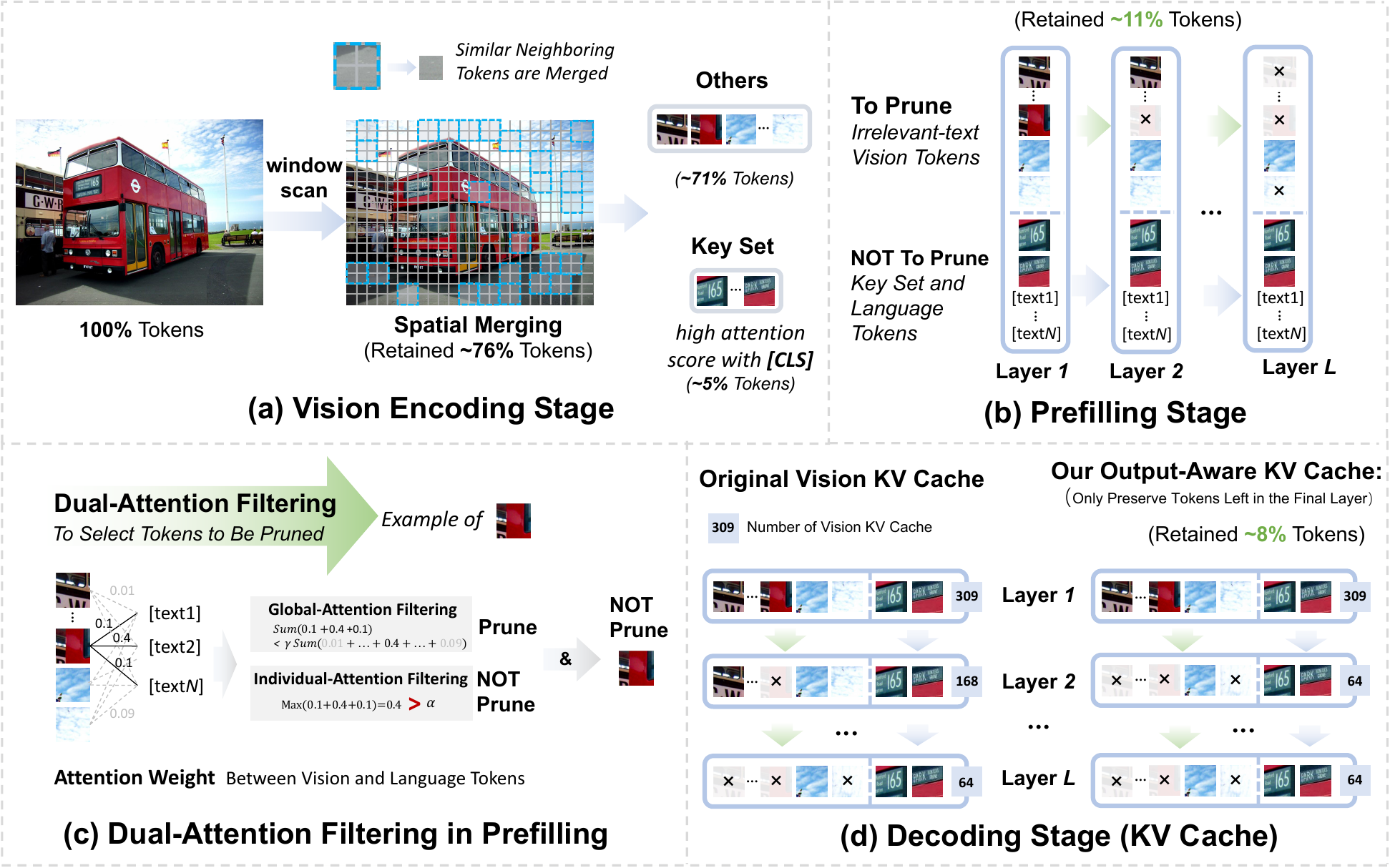}
\vspace{-6mm}
\caption{\textbf{The architecture of MustDrop.} In the vision encoding stage, MustDrop merges similar neighboring tokens by window scanning, and establishes a key set of vision-critical tokens that will not be removed at any stage. Then, the dual-attention filtering mechanism decides whether to prune tokens during prefilling. Finally, the output-aware KV cache policy further removes vision tokens during decoding.}
\vspace{-4mm}
\label{fig:overview}
\end{figure*}

\noindent\textbf{Vision Encoding:} The vision encoding stage extracts the vision information from the raw input data, usually focusing on its key regions, and extracting semantic information of objects such as their spatial positions, shapes, and relative positional relationships.
As aforementioned, vision tokens with close spatial and temporal positions exhibit highly similar and thus redundant semantic information, which are not required in the following computations, as shown in Figure~\ref{fig:overview}(a).
Besides, vision encoding can also reveal the vision tokens that are most crucial for understanding the image, which should be retained in any following stages, even if they are not relevant to any language information.

\noindent\textbf{Prefilling Stage:} The prefilling stage is usually designed to perform self-attention between all the language tokens and the vision tokens, which provides the interaction between the two modalities. Hence, the introduction of text semantics in this stage makes it possible to eliminate more redundant visual tokens that are not relevant to any language tokens. It is worth noting that such relevance should be estimated with not only the relevance to all the language tokens but also the largest relevance to any individual of the language tokens. For example, the individual text ``\textit{number}'' in Figure~\ref{fig:individual}(a) responds to the answer in the image.

\noindent\textbf{Decoding Stage}
The decoding stage is usually employed to generate language tokens based on the prefilled information. Since the understanding of the images has been accomplished in the prefilling stage, the decoding stage requires less information. As shown Figure~\ref{fig:attention map},
after the second layer, the majority of the remaining vision tokens receive almost no attention when generating the next token.
This indicates that the KV cache of vision tokens can be further compressed after the second layer.

Based on the above observations, we propose Multi-Stage Vision Token Dropping (MustDrop), a training-free token reduction method to accelerate MLLMs by identifying the redundant tokens based on their information in all three stages.
Specifically, in the vision encoding phase, MustDrop mainly eliminates the redundancy in the spatial dimension by merging the neighboring tokens with high similarity and also establishes a set of key tokens, which exhibit the most valuable vision semantic information and should not be pruned in their whole life cycle. 
In the prefilling stage, MustDrop removes the vision tokens not relevant to the language tokens gradually in multiple layers, which are selected with a dual-attention filtering strategy, estimating their global relevance to all the language tokens as well the their maximal relevance to any individual language token. In the decoding stage, after the second layer, we further compress the KV cache by pruning not only the tokens pruned in this layer but all the tokens that have been pruned in the prefilling stage.
Such a multi-stage token dropping method enables us to leverage the information from all three stages to identify the tokens that should be pruned and the tokens that should be preserved, allowing us to achieve higher performance than only using incomplete information in one of the three stages.

Extensive experiments in various benchmarks demonstrate that MustDrop can act as a plug-and-play module to improve the efficiency of MLLMs without additional training. In summary, the contributions of this work are three-fold:
\begin{itemize}[leftmargin=*,noitemsep,nolistsep]
    \item We introduce a multi-stage and training-free method to identify the tokens that should be pruned or retained in MLLMs, which considers both the vision-only information and the vision-language information, from both the prefilling stage and the decoding stage.
    \item In each stage, we introduce tailored strategies to identify the crucial tokens, such as merging the similar neighboring tokens in vision decoding, the dual attention estimation in the prefilling stage, and the output-aware cache policy in decoding stage.

    \item 
    Extensive experiments demonstrate that LLaVA \citep{liu2024visual} equipped with MustDrop achieves an efficiency and performance trade-off. For instance, MustDrop outperforms the existing state-of-the-art method SparseVLM by $ 2.1\% \sim 6.6\%$ while achieves a averaged 88.9\% compression rate on LLaVA-Next-7B.

\end{itemize}

\section{Related Work}
\label{Related Work}
\subsection{Multimodal Large Language Models}
Based on the success of large language models (LLMs) \citep{yao2024tree} such as GPTs \citep{achiam2023gpt,radford2019language}, LLaMA \citep{touvron2023llama}, and GLM \citep{glm2024chatglm} by only receiving text tokens, recent progress has been made in the area of Multimodal Large Language Models (MLLMs) \citep{li2023blip,zhu2023minigpt,wang2023cogvlm,liu2024improved} by receiving vision and text tokens. For example, LLaVA \citep{liu2024visual} encodes a 336 × 336 image into 576 tokens while also handling a relatively small number of text tokens. In contrast to text tokens, vision tokens exhibit a lower density of information and contain a substantial amount of redundancy \citep{liang2022evit}. Since the core framework of MLLM is the Transformer \citep{vaswani2017attention}
architecture, the computational cost of these models often scales quadratically with the number of input tokens. Recently, Gemini \citep{geminiteam2023gemini} and LWM \citep{liu2024world} underline the importance of long context for a comprehensive world model, and propose extending the context length to 1 million to meet growing context demands. Therefore, significantly reducing the redundancy of vision tokens while maintaining the comparable performance is an urgent issue that needs to be addressed.

\subsection{Token Compression for MLLMs} In multimodal large language models (MLLMs), visual tokens usually exceed the count of text tokens by tens to hundreds of times. In addition, vision tokens are spatial continuous and sparse in semantics compared to dense languages. Follow acceleration methods \citep{bolya2022tome,liang2022evit,liu2024sparse,zhang2024magic} in Vision Transformer (ViT), various studies \citep{shang2024llava,arif2024hired,yao2024deco} focus on removing tokens in the vision encoding phase. For example, LLaVA-PruMerge \citep{shang2024llava} proposes an Interquartile Range (IQR) method for detect important vision tokens, and reduces a large number of unimportant vision tokens at this stage. FastV \citep{chen2024image} is the first work to explore redundant visual tokens in the prefilling stage, and it drops visual tokens at the second layer during inference. A recent work, SparseVLM \citep{zhang2024sparsevlm} propose recycling the pruned visual tokens with the k highest (i.e., top-\textit{k}) values from the deleted pool. However, there are differences in the redundant tokens of different stages. They only focus on the removal of redundancy at one specific stage, and neglect the other two stages. In contrast, our work makes a more thorough study of the vision redundancy in MLLMs and proposes a multi-stage solution for vision token dropping.

\section{Methodology}
\subsection{Preliminary}
In this subsection, we explore how MLLMs sequentially process visual tokens in the three stages: vision encoding, prefilling and decoding.

\noindent
\textbf{Vision Encoding:} Vision tokens are first processed by a vision encoder. Taking the CLIP-ViT \citep{radford2021clip} in LLaVA \citep{liu2024visual} as an example, an input image is divided into a grid of patches. Each patch is converted into a token embedding by the ViT, and a learnable [CLS] token is added before these patches to compute global image information. Due to the characteristic of spatial continuity in images, there is a high probability that the surrounding tokens of a certain token will be similar. All tokens are fed into the stacked Transformer encoders, which include a multi-head self-attention (MHSA) layer and a feed-forward network (FFN).
In MHSA, input tokens are projected into three distinct matrices: $\mathbf{Q}$, $\mathbf{K}$, and $\mathbf{V}$. The attention calculates the relevance of each item to the others:
\begin{equation}
   \text{Attention}(\mathbf{Q}, \mathbf{K}, \mathbf{V}) = \text{softmax}\left(\frac{\mathbf{Q} \cdot \mathbf{K}^\mathbf{T}}{\sqrt{d_k}}\right)\cdot \mathbf{V},
\end{equation}

\noindent
where $d_k$ is the dimension of $\mathbf{K}$. The result of $\text{Softmax}(\mathbf{Q}\cdot \mathbf{K}^\mathbf{T}/\sqrt{d_k})$ is a square matrix known as the attention map.

\noindent \textbf{Prefilling:} In the prefilling stage, the text and vision tokens interact to form a unified representation, which can be denoted as $\mathbf{X}=\{\mathbf{V}_1,\mathbf{V}_2,...,\mathbf{V}_M,\mathbf{T}_1,\mathbf{T}_2,...,\mathbf{T}_N\}$, where $M$ and $N$ represent the number of image and text tokens, respectively. During the prefilling phase, the key and value of input $\mathbf{X}$ are used to construct a KV cache for each transformer layer in MLLMs. The key and value tensors are computed as follows:
\begin{equation}
\mathbf{K} = \mathbf{X} \mathbf{W_k}, \mathbf{V} = \mathbf{X} \mathbf{W_v},
\end{equation}

\noindent
where $\mathbf{W_k}$ and $\mathbf{W_v}$ are linear transformation matrices. Subsequently, $\mathbf{K}$ and $\mathbf{V}$ are stored in the KV cache to facilitate token generation during the decoding phase.

\noindent
\textbf{Decoding.} During the decoding phase, the KV cache is utilized to store the K and V for previous tokens. When generating  $\mathbf{x}_i$, only the key and value for the token $\mathbf{x}_i$ need to be computed and updated to the KV cache, while the others can be directly accessed from memory. The updates of KV cache can be formulated as

\begin{equation}
\mathbf{K} =: \text{Concat}(\mathbf{K}, \mathbf{x}_i \mathbf{W}_k),\mathbf{V} =: \text{Concat}(\mathbf{V}, \mathbf{x}_i \mathbf{W}_v).
\end{equation}

\noindent
The output of the current generated token is computed as:

\begin{equation}
    \mathbf{o_i}= \mathbf{x}_i \mathbf{W}_o, \mathbf{x}_{i}^{out}=\text{softmax}\left(\frac{\mathbf{o_i} \cdot \mathbf{K}^\mathbf{T}}{\sqrt{d_k}}\right)\cdot \mathbf{V}.
\end{equation}
 
\subsection{Token Redundancy Analysis}

In the vision encoding phase where only visual tokens are received, without the guidance of text semantics, it is difficult to determine which vision tokens are not important to the overall multimodal understanding. However, it is possible to determine the most important vision-critical tokens in this stage. These inherently significant visual information is indispensable for the entire multimodal understanding, even if it is not related to text information.

LLaVA-PruMerge \citep{shang2024llava} simply transfers the method of compress tokens in ViT will lose some text-relavent visual tokens. Adjacent pixels in an image have similar semantic relationships with each other. As shown in Figure \ref{fig:overview}(a), each object has continuous parts where spatial closeness is strongly associated with semantic similarity. There are a large number of similar tokens around some vision tokens, and these redundant tokens are not essential to occupy valuable attention during this phase and the subsequent two stages. These redundant tokens will disperse the attention of the model and affect its accuracy (See in Table \ref{tab:three-stage}). However, there are still a large number of redundant visual tokens.

\begin{figure}[t]
\centering
\includegraphics[width=1.0\linewidth]{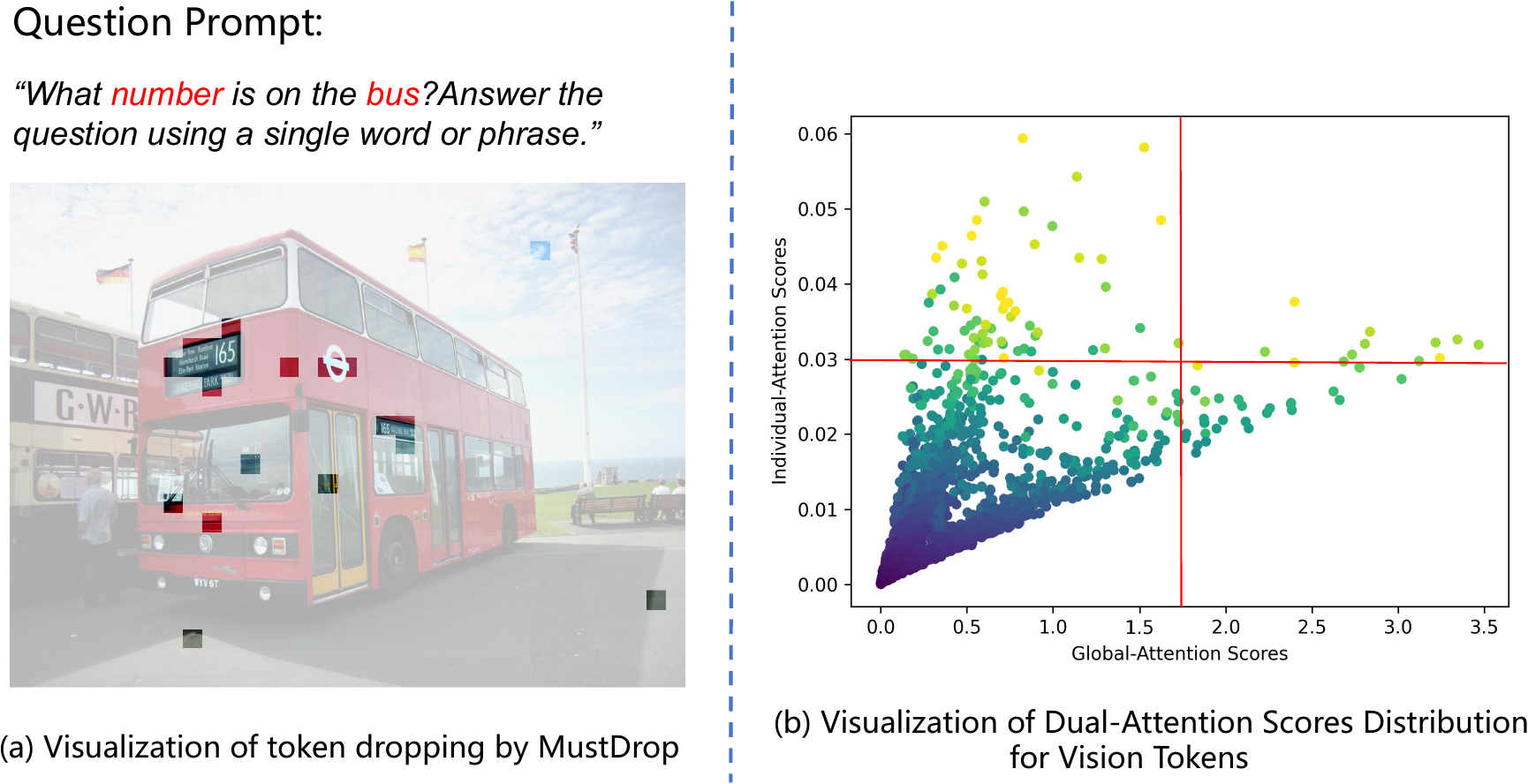}
% \vspace{-6mm}
\caption{ \textbf{The importance of individual text tokens.} (a) Visualization of token dropping by Our MustDrop. (b) Visualization of Dual-Attention Scores Distribution for Vision Tokens.}
% \vspace{-4mm}
\label{fig:individual}
\end{figure}

Subsequently, we can further remove irrelevant-text vision tokens during the prefilling stage. As shown in Figure \ref{fig:individual}, only a small fraction of text-related vision tokens actually benefit the whole multimodal understanding. In addition, individual text tokens correspond to specific region of an image. How to recognize this region guided by text is undoubtedly worth considering as a priority, to prevent it from being removed.

Given the distinct characteristics present in the above three stages of processing visual tokens, we gradually explore the feasible scheme of removing visual redundancy at each stage.

\subsection{Spatial Redundancy \& Key Recognition}
\label{Spatial Redundancy and Key Recognition}

\textbf{Local Spatial Merging.} In view of the above redundancy analysis in the vision encoding phase, we propose a Local Spatial Merging (LSM) module which does not rely on an additional network. Specifically, $M$ vision tokens can be represented as a 2D grid, which we partition into equally-sized and square windows with size $k\times k$ ($k\geq2$). Each window $w_i$ contains $k^2$ tokens, and the window represents local spatial set $w_i = \{v^i_1,v^i_2,\ldots,v^i_{k2}\}$. Subsequently, LSM calculates the similarity $v_{m,n}$ between every two different tokens for each window. The overall similarity $S_i$ of these tokens in the window $w_i$ is calculated as follows:

\begin{equation}
S_i =\sum_{\substack{m \neq n \\}} v^i_{m,n} >\tau,
\end{equation}

\noindent
where $\tau$ is a threshold. If tokens in the local window are excessively similar, LSM will merge them into a representative token by taking the weighted average. In order to prevent redundant tokens from participating in unnecessary calculations, LSM merges them in the first layer of ViT.

\noindent
\textbf{Key Retaining.} Through in-depth analysis, we have found that SparseVLM \citep{zhang2024sparsevlm} discards unimportant tokens into the pruned pool. At the same time, it computes relatively important tokens from the pool, and takes them out for subsequent reuse. This indicates that it is repeatedly computing some tokens, which is contrary to the fundamental concept of ``\textit{minimizing the computation of vision tokens as much as possible}''. Inspired by this, we look forward to identifying a set of tokens that are important enough (\textbf{Key}) at any stage. The deeper the layers in ViT, the better the [CLS] token learns the global information of visual tokens, and it no longer serves the subsequent stages. Therefore, we propose to adopt the the attention $a_{cls}$ between the [CLS] token and other tokens for identifying key vision tokens: 

\begin{equation}
    \mathbf{a_{cls}} = \text{softmax}\left(\frac{\mathbf{q_{cls}} \cdot \mathbf{K}^\mathbf{T}}{\sqrt{d_k}}\right),
\end{equation}

\noindent
where $q_{cls}$ and K represent the query vector of [CLS] and the key matrix, respectively. We set up a key set $\mathbf{O}$ to store extremely important tokens:

\begin{equation}
   \mathbf{O} = \{\mathbf{a_{cls} > \mu}\},
\end{equation}

\noindent
where $\mu$ is an adaptive threshold relied on $\mathbf{a_{cls}}$. In the subsequent two stages, namely prefilling and decoding, all outlier tokens in the set $\mathbf{O}$ will always be retained.

\definecolor{mygray}{gray}{.92}
\definecolor{ForestGreen}{RGB}{34,139,34}
\newcommand{\fg}[1]{\mathbf{\mathcolor{ForestGreen}{#1}}}
\definecolor{Forestred}{RGB}{220,50,50}
\newcommand{\fr}[1]{\mathbf{\mathcolor{Forestred}{#1}}}

\begin{table*}[h!]
  \begin{center}
    % \hspace{2mm}
    \setlength{\tabcolsep}{11.2pt}
    \renewcommand{\arraystretch}{1.4}
    \footnotesize
	\centering
	\caption{\textbf{Performance of MustDrop under different vision token configurations.} The vanilla number of vision tokens is $576$.}
    \vspace{2mm}
	\label{tab:main}
 
    \begin{tabular*}{0.79\linewidth}{c | c c c c c c c  }

        \hline
        \textbf{Method} & \textbf{GQA} & \textbf{MMB} & \textbf{MME} & \textbf{VizWiz} & \textbf{SQA} & \textbf{VQA}$^{\text{V2}}$ & \textbf{VQA}$^{\text{Text}}$\\
        \hline
        \rowcolor{mygray}
        LLaVA-1.5-7B & \multicolumn{7}{c}{\textit{Upper Bound, 576 Tokens} \ $\textbf{(100\%)}$}\\
         \textcolor{gray}{Vanilla} & \textcolor{gray}{61.9} & \textcolor{gray}{64.7} & \textcolor{gray}{1862} & \textcolor{gray}{50.0} & \textcolor{gray}{69.5} & \textcolor{gray}{78.5} & \textcolor{gray}{58.2}  \\
        
        \hline

        \rowcolor{mygray}
        LLaVA-1.5-7B & \multicolumn{7}{c}{\textit{Retain Averaged 192 Tokens} \ $\fg{(\downarrow 66.7\%)}$} \\
       ToMe \texttt{\scriptsize{(ICLR23)}} & 54.3 & 60.5 & 1563 &- & 65.2 & 68.0 & 52.1  \\
      
    FastV \texttt{\scriptsize{(ECCV24)}} & 52.7 & 61.2 & 1612 & - & 67.3 & 67.1 & 52.5 \\
    
    SparseVLM & 57.6 & 62.5 & 1721 &50.5  & 69.1 & 75.6 & 56.1 \\
    MustDrop (Ours) &  \textbf{58.2} & \textbf{62.3} & \textbf{1787} & \textbf{51.4} & \textbf{69.2} & \textbf{76.0} &  \textbf{56.5} \\
    \hline

        \rowcolor{mygray}
        LLaVA-1.5-7B & \multicolumn{7}{c}{\textit{Retain Averaged 128 Tokens} \ $\fg{(\downarrow 77.8\%)}$}\\
        ToMe \texttt{\scriptsize{(ICLR23)}}& 52.4 & 53.3 & 1343 & - & 59.6 & 63.0 & 49.1 \\
        FastV \texttt{\scriptsize{(ECCV24)}} & 49.6 & 56.1 & 1490 &-  & 60.2 & 61.8 & 50.6  \\
      SparseVLM & 56.0 & 60.0 & 1696 & 51.4 & 67.1 & 73.8 & 54.9  \\
      
      MustDrop (Ours)& \textbf{56.9} & \textbf{61.1} & \textbf{1745} & \textbf{52.1} & \textbf{68.5} & \textbf{74.6}& \textbf{56.3 }  \\
        \hline

        \rowcolor{mygray}
        LLaVA-1.5-7B & \multicolumn{7}{c}{\textit{Retain Averaged 64 Tokens} \ $\fg{(\downarrow 88.9\%)}$}\\
       ToMe \texttt{\scriptsize{(ICLR23)}} & 48.6 & 43.7 & 1138 & - & 50.0 & 57.1 & 45.3  \\

    FastV \texttt{\scriptsize{(ECCV24)}} & 46.1 & 48.0 & 1256 &- & 51.1 & 55.0 & 47.8  \\
    LLaVA-PruMerge & 48.8 & 47.4 & 1201 & 49.7 & 50.9 & 56.2 & 46.1 \\
    
       SparseVLM & 52.7 & 56.2 & 1505 & 50.1 & 62.2 & 68.2 & 51.8   \\

       MustDrop (Ours)&\textbf{ 53.1} & \textbf{60.0} & \textbf{1612} & \textbf{51.2} & \textbf{63.4} &\textbf{69.3} & \textbf{54.2}  \\

 \hline
  \hline
    \rowcolor{mygray}
        LLaVA-Next-7B & \multicolumn{7}{c}{\textit{Upper Bound, 2880  Tokens} \ $\textbf{(100\%)}$}\\
         \textcolor{gray}{Vanilla} & \textcolor{gray}{64.2} & \textcolor{gray}{67.4} & \textcolor{gray}{1851} & \textcolor{gray}{57.6} & \textcolor{gray}{70.1} & \textcolor{gray}{81.8} & \textcolor{gray}{64.9}  \\
          \hline
       \rowcolor{mygray}
        LLaVA-Next-7B & \multicolumn{7}{c}{\textit{Retain Averaged 320 Tokens} \ $\fg{(\downarrow 88.9\%)}$}\\

       SparseVLM & 56.1 & 60.6 & 1533 & 52.0 & 66.1 & 71.5 &  58.4  \\

       MustDrop (Ours)& \textbf{57.3} & \textbf{62.8} & \textbf{1641} & \textbf{54.0} & \textbf{68.0} &\textbf{73.7} & \textbf{59.9}  \\

        \hline
    \end{tabular*}
  
  \end{center}
  \vspace{-5mm}
\end{table*}

\subsection{Dual-Attention Filtering Guided by Language.}
After the coarse-grained removal of redundancy in the previous stage, there are still many redundant visual tokens. By the guidance of text semantics in the prefilling stage, we design dual-attention filtering mechanism to evaluate the importance of the remaining visual tokens. 

\noindent
\textbf{Global-Attention Filtering.}
It is necessary to understand the correlation between vision token $v_{j}$ and global text semantics. During the prefilling phase, the attention score $a_{i,j}$ between $v_{j}$ and a text token $t_{i}$ has already been calculated. Inspired by this, we adopt the attention mechanism to assist MustDrop in identifying unimportant vision tokens. We can get the global attention $V_{j}$ between vision token $v_{j}$ and all text tokens in a very efficient manner as follows:

\begin{equation}
V_{j} =\sum_{i=1}^{N} a_{i,j}. 
\end{equation}

\noindent
Then, we can dynamically ascertain the set $D$ of vision tokens, that exhibit the least correlation with the textual content by the threshold $\gamma$:

\begin{equation}
D= \{V_j\leq \gamma \sum_{j=1}^{M} V_{j}\},
\end{equation}

\noindent
note that the number $d$ of vision tokens in $D$ is dynamically changing, rather than fixed.

\begin{figure}[t]
\centering
\includegraphics[width=0.9\columnwidth]{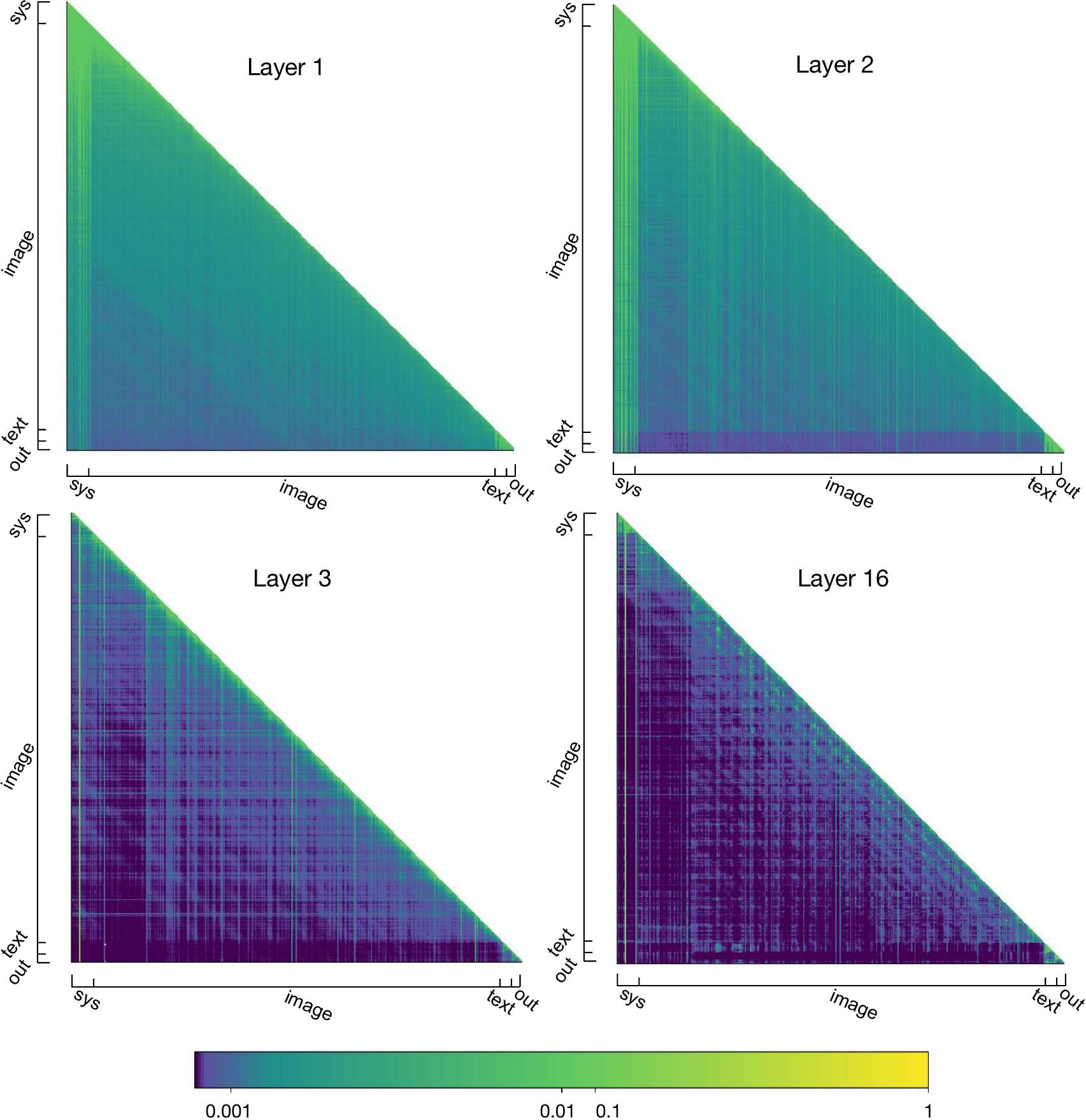}
\caption{Visualization of attention during the decoding process of for LLaVA1.5-7B at certain layers. The attention maps of all layers can be seen in the Appendix.
}
\label{fig:attention map}
\end{figure}

\noindent
\textbf{Individual-Attention Filtering.}
The set $D$ determined by the global-attention filtering policy may contain a very small number of tokens that are falsely considered unimportant. As shown in Figure \ref{fig:individual}(a), some words are irrelevant to the image (e.g., prepositions and pronouns) while individual word such as ``number'' to the corresponding region of the image. This indicates that some individual text tokens are highly significant and may be overshadowed by the global textual semantics. Therefore, individual estimation is proposed to address the issue, which is crucial for maintaining performance (See in Table \ref{Table:dual-attention}). We assess the correlation between each text token and the visual token $j$. The calculation for individual estimation is as follows:

\begin{equation}
\max_{j} a_{ji} < \alpha.
\end{equation}

\noindent
In the case where the maximum attention value between visual token $j$ and individual text token is still relatively small, this visual token $j$ is considered to be truly unimportant. The $\alpha$ is a threshold.

As seen in Figure \ref{fig:individual}(b), there exist some vision tokens whose global-attention scores are low while their individual-attention scores are high, and the opposite situation also exists. Thus, our dual-attention filtering mechanism is a relatively comprehensive assessment strategy.

\noindent
\subsection{Output-aware Cache Policy}
In the decoding process of MLLMs, the design of the KV cache is intended to accelerate the speed of generating output tokens. However, it also imposes a huge burden on storage. Although we have dropped a large number of visual tokens in the previous vision encoding and prefilling stages, there is still a significant storage cost for KV cache. Therefore, we visualize and analyze the attention maps during the decoding process for LLaVA. As shown in Figure \ref{fig:attention map}, it can be observed that following the second layer, the attention existing between the image tokens and the output tokens is extremely sparse. Considering that the prefilling stage undergoes multiple pruning for vision tokens in different layer, the set $S_{few}$ of visual tokens still retained in the last layer are the fewest, and they play a crucial role in the entire multimodal understanding. Motivated by this, we propose an output-aware cache policy. 
During the decoding phase, we only store the vision tokens that belong to set $S_{few}$ into KV cache starting from the shallow layers.
The policy further drops the KV cahce of vision tokens in the decoding stage.

\begin{table*}[!h]
\centering
% \vspace{-0.1in}
\caption{A comparison of different MLLMs on video-based multimodal reasoning benchmarks. The original number of video tokens is 2048, FastV and our Mustdrop retain an average of 50\% tokens. The GPT-3.5 turbo is adopted for assistive evaluation.} 
% \vspace{-0.1in}
\scalebox{0.8}{
\begin{tabular}{@{}lc|cc|cc|cc|cc@{}}
\toprule
\multirow{2}{*}{Methods}         & \multirow{2}{*
}{LLM size}  & \multicolumn{2}{c|}{\textbf{TGIF}}& \multicolumn{2}{c|}{\textbf{MSVD}} & \multicolumn{2}{c|}{\textbf{MSRVT}}  & \multicolumn{2}{c}{\textbf{Avg.}}  \\ 
                &            & Accuracy    &Score             
& Accuracy    & Score        & Accuracy     & Score         & Accuracy    & Score             \\ \hline
FrozenBiLM \cite{yang2022zero}     & 1B         & 41.9         &-                 
& 32.2        & -            & 16.8         & -             &30.3 &- \\
VideoChat \cite{li2023videochat}     & 7B         & 34.4            &2.3              
& 56.3        & 2.8          & 45.0         & 2.5           &45.1 &2.5 \\
LLaMA-Adapter   & 7B         & -         &-               
& 54.9        & 3.1          & 43.8         & 2.7           &- & -
\\
Video-LLaMA \cite{zhang2024llama}   & 7B         & -         &-              
& 51.6        & 2.5          & 29.6         & 1.8           & -& -\\
Video-ChatGPT \cite{maaz2023video}  & 7B         & 51.4         &3.0               
& 64.9        & 3.3          & 49.3         & 2.8           &55.2 & 3.0
\\
\hline
\hline
 
Video-LLaVA \cite{lin2023video}    & 7B         & 47.0 &3.4
& 70.2 & 3.9 & 57.3 & 3.5  & 58.2&3.6 \\
\hline
+ FastV \cite{chen2024image}  & 7B         &45.2 &3.1
& 71.0 &\textbf{3.9} &  55  & 3.5   & 57.1 &3.5 \\

% \rowcolor{mygray}
+ MustDrop (Ours) & 7B     &\textbf{46.2} &\textbf{3.3}&\textbf{71.5} &\textbf{ 3.9} & \textbf{56.5}  &\textbf{3.6} & \textbf{58.1} & \textbf{3.6} \\ \bottomrule

\end{tabular}}
% \vspace{-0.2in}
\label{tab:main_table_video}
\end{table*}
\section{Experiments}

\subsection{Evaluation Tasks}
We validate MustDrop on a wide range of multimodal benchmarks to assess its effectiveness, including image-based and video-based multimodal understanding tasks. For image-based multimodal evaluation, we conduct experiments on seven widely adopted benchmarks including GQA, \citep{hudson2019gqa}, MMBench (MMB) \citep{liu2025mmbench}, MME \citep{fu2023mme}, VizWiz \citep{bigham2010vizwiz}, SQA \citep{lu2022learn}, VQA$^{\text{V2}}$ (VQA V2) \citep{goyal2017making}, and VQA$^{\text{Text}}$ (TextVQA) \citep{singh2019towards}. Furthermore, We test MustDrop on three video-based multimodal understanding tasks, TGIF-QA \citep{jang2017tgif}, MSVD-QA \citep{xu2017video}, and MSRVTT-QA \citep{xu2017video}. More details are included in the Appendix.

\subsection{Implementation Details}
We test MustDrop with various open-source MLLMs. For image-based multimodal understanding tasks, we conduct experiments not only on LLaVA-1.5-7B \citep{liu2024visual} frameworks, but also on LLaVA-Next-7B \citep{liu2024llavanext} for further validating high-resolution images. They both adopt CLIP-ViT-L as the visual encoder. When it comes to video-based multimodal understanding tasks, our baseline model is Video-LLaVA \citep{lin2023video}. For a fair comparison, we adopt the settings as reported in their paper for the baseline models. More implementation details are included in the Appendix.

\subsection{Main Results}

\textbf{Image-based multimodal understanding.} In table \ref{tab:main},
we verify on LLaVA-1.5-7B and LLaVA-Next-7B with seven representative datasets, respectively. For LLaVA-1.5-7B, MustDrop compares with mainstream methods at different sparsity levels, i.e., an average of 192, 128 and 64 retained tokens. Unsurprisingly, ToMe and LLaVA-PruMerge directly merge visual tokens only in vision encoding without considering text information, resulting in unideal results. Even though FastV and SparseVLM have both considered textual guidance to remove vision tokens individually, they still do not demonstrate superiority over MustDrop when retaining the same average number of tokens. The reason is that we thoroughly analyze the characteristics of visual semantics in the vision encoding, prefilling, and decoding stages, respectively. Thanks to the tailored design of Local Spatial Merging (LSM) module, dual-attention filtering mechanism and output-aware cache policy, our MustDrop enables redundant visual tokens to be precisely dropped at each stage. To verify the effectiveness of MustDrop on high-resolution images, we also conduct verification on the LLaVA-Next-7B model. Without loss of generality, when an average of 88.9\% tokens are removed in both cases, our method outperforms the existing SOTA method SparseVLM on all datasets.

\noindent
\textbf{Video-based multimodal understanding.} To assess the generalization capabilities of MustDrop across different modalities, we further extend it to Video-LLaVA. Video-LLaVA samples 8 frames from a video and extracts 2048 vision tokens. For fair comparison, we retain an average of 50\% vision tokens followed by FastV \citep{chen2024image}. It is obviously that MustDrop outperforms FastV across all benchmarks, both in accuracy and evaluation score. Note that compared to the original Video-LLaVA, the performance of Mustdrop is lossless and even shows an 1.3\% accuracy improvement on MSVD dataset. This demonstrates that MustDrop has strong generalization capabilities in the video modality. While using only about half of vision tokens, it generates accurate responses to a variety of questions.

\subsection{Ablation Study}

\renewcommand{\arraystretch}{0.7}
\begin{table}[!t]
  \caption{\textbf{Ablation study of specific strategy at vision encoding, prefilling, and decoding stage, respectively.} The detailed metric includes the numbers of vision tokens, storage (KV cache memory), and performance (Acc.). Note the number of vision tokens refers to that in the KV cache.}
  \renewcommand{\arraystretch}{1}
  \label{tab:three-stage}
  \centering
  \small
  \scalebox{0.85}{
  
  \begin{tabular}{lccc}
    \toprule
    \multirow{2}{*}{\textbf{Method}} & \multirow{2}{*}{\textbf{Tokens}} &  \hspace{2mm}\textbf{VQA}$^{\text{Text}}$ & \textbf{KV Cache}  \\
     & & \textbf{Acc.} & \textbf{Memory (MB)}  \\
    \midrule
    Baseline & 576 & 58.2 & 302.4 \\
    \midrule
     
     \hspace{3mm}\textbf{Vision Encoding} & \multirow{2}{*}{Avg. 440} & \multirow{2}{*}{\textbf{58.3}} & \multirow{2}{*}{231.0 {\color[HTML]{228B22}($\downarrow 23.6\%$)}} \\
     Local Spatial Merging  \\
      \hline
    \hspace{8mm}\textbf{Prefilling}  & \multirow{2}{*}{Avg. 64} & \multirow{2}{*}{55.1} &  \multirow{2}{*}{33.6 {\color[HTML]{228B22}($\downarrow 88.9\%$)}} \\
    Dual-Attention Filtering \\
      \hline
    \hspace{8mm}\textbf{Decoding} & \multirow{2}{*}{\textbf{Avg. 45}} & \multirow{2}{*}{54.2} & \multirow{2}{*}{\textbf{23.6} {\color[HTML]{228B22}($\downarrow 92.2\%$)}} \\
    Output-Aware Cache  \\
    
    \bottomrule
  \end{tabular}
  }
  \vspace{-4pt} 
\end{table}

\textbf{Stage-Aware Strategy.} We conduct an ablation experiment on TextVQA to examine the benefits and the impact on performance brought by the specific dropping strategies at each stage. From Table \ref{tab:three-stage}, we can see that: \textbf{(1)} In the vision encoding stage, Local Spacial Merging (LSM) module eliminates redundant tokens by merging the neighboring tokens with high similarity. To our surprise, LSM could improve the performance while dropping averaged 136 vision tokens and reducing 23.6\% KV cache memory; \textbf{(2)} In the prefilling stage, we design the dual-attention filtering mechanism between text and vision tokens to largely prune visual tokens unrelated to the text semantics. Under the case of further reducing an average of 376 vision tokens and reducing 65.3\% KV cache memory, the performance has only dropped by 3.2\%; \textbf{(3)} In the decoding stage, output-aware policy is proposed to prune irrelevant-output vision tokens in shallower layers. Compared to the prefilling stage, vision KV cache memory is further reduced within a smaller decline of performance. Through an in-depth analysis of the redundant nature of visual tokens at each stage, the tailored strategies of MustDrop at each specific stage are capable of precisely removing invalid visual tokens.

\noindent
\textbf{Dual-Attention Filtering Mechanism.} To understand the effectiveness of the dual-attention mechanism, we evaluate two datasets (i.e.,TextVQA and MMB) in Table \ref{Table:dual-attention}. The results indicate that: \textbf{(a)} In this setup, we retain an average of 64 vision tokens based on their correlation with the [CLS] token in the vision encoder. Without the guidance of textual information, removing a large number of vision tokens will undoubtedly damage the performance to the greatest extent; \textbf{(b)} Only introducing the global-attention between vision and language, the average performance across the three datasets improve by 4.7\%; \textbf{(c)} Compared to (b), only introducing individual-attention strategy brings similar benefits, with an average performance of 52.9\%; \textbf{(d)} Considering both global and individual attention filtering strategies simultaneously can maximize performance improvements. Compared to (a), there is a 5.6\% increase in performance. The experimental results and visualization in Figure \ref{fig:individual}(b) have reached a consistent conclusion: relying solely on the global or local aspects will lead to the loss of important semantic information. Therefore, our dual-attention filtering mechanism is relatively comprehensive.

\begin{table}[!t]
\centering
\caption{\textbf{Effectiveness of dual-attention filtering Mechanism} in the prefilling stage. The average number of tokens retained is 64 for all cases. Note the ablation study removes the Key Retaining mechanism.}
\small
\setlength{\tabcolsep}{6pt}

\scalebox{1}{\begin{tabular}{l|c|c|cc|c}
\toprule
    
 \multirow{2}{*}{\#} & \multirow{2}{*}{\textbf{Global}} & \multirow{2}{*}{\textbf{Individual}} & \multicolumn{2}{c|}{\textbf{Performance}} & \multirow{2}{*}{\textbf{Avg.}}\\

 & &  & VQA$^{\text{Text}}$ & MMB  \\ \midrule

(a)  &  & & 47.3  & 50.2&48.8 \\
\addlinespace[2pt]

(b)  & \checkmark & &  52.2 & 54.7 & 53.5 \\
\addlinespace[2pt]

(c) & & \checkmark  & 51.2 & 54.5  & 52.9 \\
\addlinespace[2pt]
\rowcolor{gray!20}
(d) & \checkmark& \checkmark  & \textbf{52.8} & \textbf{56.0}  & \textbf{54.4}\\

\bottomrule

\end{tabular}
}
\vspace{-4mm}

% \vspace{-4mm}
\label{Table:dual-attention}
\end{table}

\noindent
\textbf{Key Retaining Mechanism.}
To verify that some visual tokens are inherently important, we study the importance of the key vision tokens recognized in vision encoder. As shown in Table \ref{tab:key set}, Key Retaining (KR) mechanism achieves a significant average performance improvement of 1.5\% and 3.0\% on TextVQA and MMB, respectively. This indicates that recognizing naturally important visual tokens is crucial during the vision encoder phase.

\begin{table}[t] 
    \centering
    \setlength{\tabcolsep}{6pt}
    \renewcommand{\arraystretch}{1.4}
    \footnotesize
	\caption{\textbf{Ablation study on key retaining (KR).} Experiments are conducted on TextVQA and MMB using LLaVA.}
 \vspace{2mm}
	\label{tab:key set}
     \begin{tabular}
     {cc c  c }
     % \centering
     \toprule
     \multirow{2}{*}{\textbf{Benchmark}} & \multicolumn{2}{c}{\textbf{Tokens}} & \multirow{2}{*}{\textbf{Avg.}}\\ 
    \cline{2-3}
    ~ & \textbf{64} & \multicolumn{1}{c}{\textbf{128}}  & \\
    \midrule
    \text{Text} VQA & 52.8&\multicolumn{1}{c}{54.8}& 53.8 \\
    \rowcolor{mygray}
    \textbf{+} KR & \textbf{54.2}$(\fr{\uparrow1.4})$&
    \multicolumn{1}{c}{\textbf{56.3}$(\fr{\uparrow1.5})$}&
   
    \textbf{55.3}$(\fr{\uparrow1.5})$  \\
    \midrule
    
    MMB &56.0 &\multicolumn{1}{c}{59.2}&  57.6 \\
    \rowcolor{mygray}
    \textbf{+} KR & \textbf{60.0}$(\fr{\uparrow4.0})$&
    \multicolumn{1}{c}{\textbf{61.1}$(\fr{\uparrow1.9})$}&

    \textbf{60.6}$(\fr{\uparrow3.0})$    \\
    \bottomrule
     \end{tabular} 
     \vspace{-2mm}
\end{table}

\subsection{Efficiency Analysis}
MustDrop affords significant efficiency and storage gains during the inference process. We compare the efficiency indicators of Tokens, Total-Time, FLOPs and GPU-Memory with vanilla LLaVA-Next-7B and SparseVLM. We achieve a dramatic reduction in the number of visual tokens, decreasing the original count by approximately 88.9\% in all datasets. Taking the TextVQA dataset as an example, we can observe from Table \ref{tab:efficiency} that within a reasonable range of performance degradation, MustDrop achieves remarkable efficiency. Concretely, only approximately 11.1\% of the tokens are used, which significantly reduces the data volume. Benefiting from this, the speed is increased by about 41.5\%. Moreover, the storage space is decreased by 1344.1 MB, and the FLOPs are reduced by about 88.5\%. As analyzed in Subsection \ref{Spatial Redundancy and Key Recognition}, the strategy of SparseVLM for recognizing redundant tokens is overly complicated, and the computational cost brought by its strategy is higher than the cost saved by removing redundant tokens. Therefore, while retaining the same number of tokens, its inference time is slower 15.6\% than our MustDrop.

\begin{table}[!t]
    \centering
    \caption{Inference costs of the number of tokens, Total-Time, FLOPs, and KV Cache Memory using LLaVA-Next-7B.}
        \setlength{\tabcolsep}{3.5pt}
            \renewcommand{\arraystretch}{1.2}
            \footnotesize
    \begin{tabular}{cccccc}
        \toprule
         \multirow{2}{*}{Methods}   & \multirow{2}{*}{Tokens} & Total-Time & FLOPs  & KV Cache  & VQA$^{\text{Text}}$ \\
           && (Min \& Sec) & (T) & (MB)  & Acc. \\
         \midrule
         Vanilla & 2880 & 26:34 & 9.6  &1512.1 &\textbf{64.9} \\
         \hline
         SparseVLM & \textbf{320}& 18:26 &1.5& \textbf{168.0}&58.4 \\
         MustDrop & \textbf{320} &\textbf{15:33} & \textbf{1.1} & \textbf{168.0}&59.9 \\
       
        \bottomrule
    \end{tabular}
    \label{tab:efficiency}
\end{table}

% \begin{table}[!t]
%     \centering
%     \caption{Inference costs of the number of tokens, Total-Time, FLOPs, and KV Cache Memory using LLaVA-Next-7B.}
%         \setlength{\tabcolsep}{2pt}
%             \renewcommand{\arraystretch}{1.2}
%             \footnotesize
%     \begin{tabular}{ccccccc}
%         \toprule
%          \multirow{2}{*}{Methods}   & \multirow{2}{*}{Tokens} & Total-Time & FLOPs  &  Cache & Latency & VQA$^{\text{T}}$ \\
%            && (Min \& Sec) & (T) & (MB) & (ms) & Acc. \\
%          \midrule
%          Vanilla & 2880 & 26:34 & 9.6  &1512.1 &&\textbf{64.9} \\
%          \hline
%          SparseVLM & \textbf{320}& 18:26 &1.5& \textbf{168.0}&23&58.4 \\
%          MustDrop & \textbf{320} &\textbf{15:33} & \textbf{1.1} & \textbf{168.0}&\textbf{20}&59.9 \\
       
%         \bottomrule
%     \end{tabular}
%     \label{tab:inference}
% \end{table}

\section{Conclusion}
\label{con}
In this paper, we improve the efficiency of  Multimodal Large Language Models (MLLMs) from the perspective of dropping the number of visual tokens. We propose MustDrop, a plug-and-play inference cost optimization method for MLLMs. Unlike existing methods, MustDrop measures the importance of each vision token from the whole lifecycle, i.e., the vision encoding stage, prefilling stage, and decoding stage. MustDrop significantly enhances the efficiency of MLLMs in image and video understanding tasks. Our method exceeds the existing SOTA SparseVLM by $ 2.1\% \sim 6.6\%$ while achieves a averaged 88.9\% compression rate on LLaVA-Next-7B.

\clearpage
{
    \small
    \bibliographystyle{ieeenat_fullname}
    \bibliography{ref}
}


\end{document}